\documentclass{article}
\usepackage[preprint,nonatbib]{neurips_2026}
\usepackage[authoryear,round]{natbib}

\usepackage{graphicx}
\graphicspath{{figures/}}
\usepackage{amsmath,amssymb}
\usepackage{array}
\usepackage[ruled,lined,linesnumbered,longend]{algorithm2e}
\usepackage{placeins}
\usepackage{fvextra}
\usepackage{xcolor}
\usepackage[colorlinks=true,linkcolor=blue!55!black,citecolor=blue!55!black,urlcolor=blue!55!black,bookmarksnumbered=true,pdfusetitle]{hyperref}
\newcommand{\LTwoDUniform}{L2D Uniform}
\newcommand{\LTwoDClusteredMixed}{L2D C/M}

\usepackage{xcolor}

\title{Revisiting Learning to Delegate: LLM Discovered Heuristics for Large-Scale Vehicle Routing}
\author{%
  \normalfont
  \begin{tabular}[t]{@{}l@{\hspace{3em}}l@{}}
    \textbf{Zhengyang Wang}\thanks{Equal Contribution.} &
    \textbf{Renfei Tan}\footnotemark[1]\\
    University of Michigan & Massachusetts Institute of Technology\\
    \texttt{zmaxwang@umich.edu} & \texttt{rftan@mit.edu}\\[1em]
    \textbf{Shuze Daniel Liu} &
    \textbf{David Simchi-Levi}\thanks{Corresponding author.}\\
    Massachusetts Institute of Technology & Massachusetts Institute of Technology\\
    Purdue University & Purdue University\\
    \texttt{shuzel@mit.edu} & \texttt{dslevi@mit.edu}
  \end{tabular}
}
\date{November 2026}
\hypersetup{pdfauthor={Zhengyang Wang, Renfei Tan, Shuze Daniel Liu, David Simchi-Levi}}

\begin{document}
\maketitle

\begin{abstract}
Efficient vehicle routing is critical to logistics and transportation, yet solving large-scale capacitated vehicle routing problems (CVRP) remains computationally challenging due to their combinatorial complexity. One approach to solve large-scale vehicle routing optimization is to identify promising subproblems and then optimize them using solvers.
In this paper, we propose a framework of automatic heuristic design (AHD) in which a large language model (LLM) autonomously refines heuristics for selecting subproblems, enabling effective optimization of large-scale CVRP instances. 
Compared to prior approaches that train task-specific neural networks to select among heuristically constructed subproblems, our method discovers generalizable, interpretable LLM-generated heuristics to directly construct subproblems.
Experiments show that our approach achieves comparable in-distribution performance while outperforming neural network-based policies on out-of-distribution instances with large-scale problems.
\end{abstract}

\section{Introduction}
\label{sec:introduction}

The vehicle routing problem (VRP) has remained a classical combinatorial optimization problem with both theoretical interest and ample applications for decades.
Many solvers for large-scale vehicle routing problems use hand-crafted heuristics whose performance is assessed empirically. 
Developing these heuristics often involves repeatedly designing and testing a variety of rules across a wide range of VRP instances, checking both the cost of routing and running time.

Large language models (LLMs) offer a way to automate some parts of this process.
They can generate heuristics \citep{liu2024eoh}, receive feedback, and use that feedback to propose new heuristics. However, designing a complete VRP solver from scratch involves many components, and evaluating these solvers can be computationally expensive on large instances.

We therefore revisit the select-solve-merge framework of Learning to Delegate (L2D; \citealp{li2021learning}) to give the language model a specific design task and evaluate its proposals under fixed time budgets. Within this framework, the model discovers the heuristic that selects subproblems for reoptimization.
Given a feasible VRP solution, the generated heuristic constructs and prioritizes groups of current routes. Each group defines a subproblem, also called a “delegation,” whose customers are jointly reoptimized by a fixed repair solver. Improving repairs are merged back into the current solution, and the process repeats. The overall framework and repair solver remain fixed during discovery. 
In this paper, we ask whether a language model can discover an effective delegation heuristic, and whether the selected heuristic remains useful when instance distributions, problem sizes and repair solvers change.

In summary, our contributions are as follows:
\begin{itemize}
    \item We develop a hypothesis-driven approach to automatic heuristic design (AHD) within L2D's select-solve-merge framework. An LLM proposes and refines programs for building and ranking delegations.
    \item The primary heuristic program achieves comparable cost reductions to the L2D models on instances drawn from their training distributions. On instances from new distributions, it achieves greater reductions than both models in majority of the tested size.
    \item In an exploratory comparison, two other discovered programs during AHD achieved higher observed mean reductions than the selected program in every test setting, despite ranking below the primary program on validation.
\end{itemize}
\section{Related work}
\label{sec:related-work}

\subsection{Vehicle routing methods}

Many heuristics for large vehicle routing problems improve an existing solution by reoptimizing part of it at a time.
In Shaw's large neighborhood search, selected customer visits are removed and reinserted using constraint programming~\citep{shaw1998using}.
POPMUSIC repeatedly reoptimizes groups of related parts of a solution~\citep{taillard2002popmusic}.
In experiments with two heuristics for the capacitated vehicle routing problem (CVRP), \citet{santini2023decomposition} found that grouping whole routes generally produced better solutions than merging short sequences of customer visits into single stops.

Learned models can guide both the construction and improvement of routes.
\citet{kool2019attention} train a model to construct routing solutions, and \citet{hottung2020neural} train a model to rebuild part of a solution during large neighborhood search.
GLOP uses learned models to partition large instances and construct routes within those partitions~\citep{ye2024glop}.
\citet{bi2022amdkd} use knowledge distillation to train a routing model using the outputs of models trained on different instance distributions.
\citet{yu2026coeks} let components that specialize in different routing constraints learn from one another's outputs to help the model handle unseen combinations of constraints.

Learning to Delegate (L2D) uses learning to decide which routes to reoptimize together~\citep{li2021learning}.
It represents each route by the average location of its customers and uses nearby route centers to form candidate groups.
It trains a neural network to predict the routing costs returned by the repair solver for these groups, then selects the group with the largest predicted cost reduction.

\subsection{Automatic heuristic design}

In LLM-based automatic heuristic design (AHD), the model proposes heuristic programs and uses evaluation feedback to guide later proposals.
At each iteration, FunSearch samples existing programs from its database and uses their source code to prompt an LLM to generate new programs~\citep{romeraparedes2024funsearch}.
EoH jointly evolves natural-language descriptions of heuristic ideas and their implementations~\citep{liu2024eoh}.
ReEvo uses an LLM to compare pairs of heuristics and explain their performance differences. It accumulates lessons from these comparisons across iterations to decide how later programs are combined and modified~\citep{ye2024reevo}.

AHD has also been studied across optimization problems and classes of instances.
CO-Bench evaluates agents that search for algorithms across 36 combinatorial optimization problems~\citep{sun2025cobench}.
LACE first specifies the required form of a heuristic program and generates initial heuristics. It then evolves a collection whose members perform well on different instances within runtime limits~\citep{gong2026lace}. MTHS searches over descriptions of general heuristic-search procedures and refines programs that implement them for particular problems~\citep{liu2026mths}.
\citet{baek2026llms} studies algorithms intended to work across classes of instances in inventory control, queueing control and assortment optimization.

In vehicle routing, AHD can redesign individual decisions within an existing solver.
SpecAHD~\citep{lai2026specahd} searches for programs that select parts of a solution to repair and for heuristics that rebuild those parts.
AILS-AHD evolves a rule that chooses which customers to remove from the current solution~\citep{xie2026ailsahd}.
VRPAgent jointly evolves rules for choosing which customers to remove and the order in which to reinsert them during large neighborhood search~\citep{hottung2026vrpagent}.
PyVRP+ uses design hypotheses to explain why proposed changes might improve search.
It evolves rules for choosing and retaining candidate solutions and updating penalties for constraint violations in hybrid genetic search~\citep{malik2026pyvrpplus}.
ARS generates code that adapts an existing heuristic search to the problem's constraints~\citep{li2025ars}.

\section{Methodology}
\label{sec:methodology}

\subsection{Delegation}

In the capacitated vehicle routing problem (CVRP), a feasible solution serves every customer exactly once on depot-to-depot routes that respect vehicle capacity. 
Its cost is the sum of continuous Euclidean edge lengths. 

Following L2D~\citep{li2021learning}, we iteratively select a \emph{delegation}, a group of whole routes from the current solution. The customers on those routes define a routing subproblem. The repair solver reoptimizes the routes for those customers while keeping the other routes fixed. When a feasible repair lowers cost, we merge it into the current solution and make the next selection from the updated routes.

L2D and our method differ in how they form delegations and decide which to try first.
\begin{itemize}
  \item \textbf{L2D} uses a fixed rule to group each route with those whose centers are closest to its own. A trained neural network ranks the resulting groups by predicted cost reduction.
  \item \textbf{Our AHD} aims to produce a single program that constructs delegations and returns them in the order the procedure should try them.
\end{itemize}

We use either the Lin--Kernighan--Helsgaun solver (LKH)~\citep{helsgaun2017lkh3} or hybrid genetic search (HGS)~\citep{vidal2012hybrid,vidal2022hgs} to repair the selected delegation.

\begin{figure}[t]
  \centering
  \includegraphics[width=\linewidth]{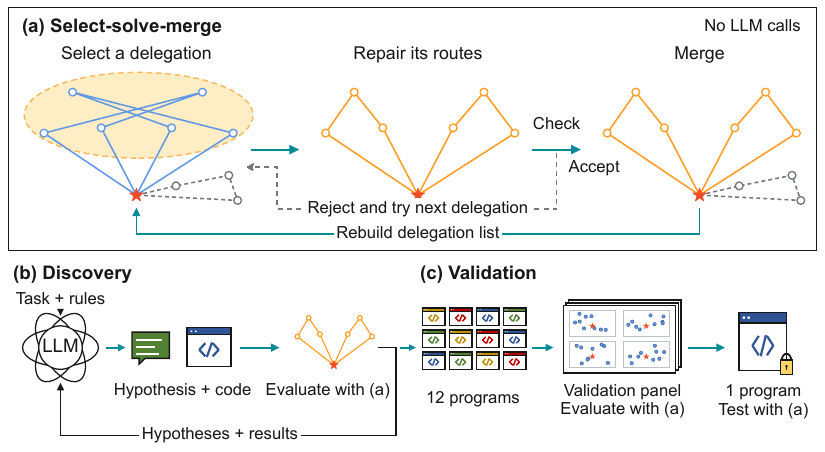}
  \caption{Select-solve-merge procedure, program discovery, and validation.
  (a) \textit{Select:} the program forms and ranks among a list of delegations, each group of current routes defining a routing subproblem over their customers. \textit{Solve:} a fixed solver reoptimizes the routes for the customers in the selected delegation. \textit{Merge:} the procedure merges a feasible improvement and rebuilds the list, or continues through the current list.
  (b) The LLM proposes a design hypothesis and a program whose evaluation guides later proposals.
  (c) A separate validation set is used to choose one shortlisted program for testing.}
  \label{fig:autoresearch-pipeline}
\end{figure}

Figure~\ref{fig:autoresearch-pipeline} summarizes the select-solve-merge procedure, program discovery, and validation.

Let $P$ denote a routing instance and $R$ the history of completed repair attempts in the current run. The routing state $X$ contains the current feasible routes, their total cost, and how many times each customer has been selected for repair. The limits $\ell$ bound the number of routes and customers in each delegation and the length of the returned list. A program $h$ returns an ordered list of $K$ delegations.
\begin{equation}
 (S_1,\ldots,S_K)\leftarrow h(P,X,R,\ell).
 \label{eq:delegation-list}
\end{equation}

Each $S_j$ is a group of whole current routes, represented by their route IDs. Algorithm~\ref{alg:delegation} processes these groups in the returned order. Appendix~\ref{app:program-interface} gives the program interface and its limits.

A delegation is eligible if its exact customer set has yet to be used in a completed repair attempt in the current run. Completed attempts count whether the repair is accepted or rejected. We compare customer sets because route IDs can change after a repair.
Costs from separate runs of the solver on the full instance, distribution labels and direct solver calls are unavailable to the program.

The routing budget includes program execution, solving, merging and feasibility and cost checks. Only improvements completed and checked within this budget are accepted.

\begin{algorithm}[t]
\SetAlgoLined
\KwIn{Problem $P$, routing state $X$, program $h$, limits $\ell$, repair solver, routing budget $T$}
$R\leftarrow\varnothing$; $\mathcal C\leftarrow h(P,X,R,\ell)$\;
\While{elapsed time $<T$ and $\mathcal C$ has an eligible delegation}{
    $S\leftarrow$ next eligible delegation of $\mathcal C$\;
    Attempt to reoptimize the customers in $S$; keep routes outside $S$ fixed\;
    Check the returned routes; record any completed solve in $R$ and update selection statistics in $X$\;
    \eIf{a feasible strict cost reduction is completed and checked before $T$}{
        Replace the affected routes and cost in $X$\;
        $\mathcal C\leftarrow h(P,X,R,\ell)$\;
    }{
        Retain the routes in $X$ and advance the existing list\;
    }
}
\KwOut{Current feasible routes in $X$}
\caption{Iterative select-solve-merge procedure}
\label{alg:delegation}
\end{algorithm}

\FloatBarrier

\subsection{Objective}
We score programs by their \emph{mean cost reduction} and use \emph{paired advantage} to compare them with baselines on the same instances.

\paragraph{Mean cost reduction.}
On each instance, reduction is the percentage of the starting routing cost saved by the end of the routing budget. 
During discovery, we maximize the average reduction over the instances in the discovery panel $D$, the common set of instances used to evaluate proposals.
For instance $i$, let $C_i^0$ be the cost of its supplied feasible starting solution and $C_i(h,T,s)$ the final cost returned using program $h$, routing budget $T$ and solver seed $s$.
\begin{equation}
 \operatorname{Eval}(h,D,T,s)=\frac{1}{|D|}\sum_{i\in D}100\frac{C_i^0-C_i(h,T,s)}{C_i^0}.
 \label{eq:eval-reduction}
\end{equation}
Appendix~\ref{app:objective-choice} explains how normalization affects the comparison.

\paragraph{Paired advantage.}
We compare each generated program with the three baselines described in Section~\ref{sec:test-panels}. Paired advantage is the mean difference in reduction between a program and a baseline on the same instances.
Each test panel pairs a set of instances with one repair solver.

For any evaluated method $a$, let $C_{ias}(t)$ be its solution cost on instance $i$ at elapsed time $t\leq T$ with seed $s$. For validation and testing, we average costs over three solver seeds within each instance, then average the reductions over the $n$ instances in the panel.
\begin{equation}
 \bar C_{ia}(t)=\frac13\sum_{s=0}^2C_{ias}(t),\qquad
 r_{ia}(t)=100\frac{C_i^0-\bar C_{ia}(t)}{C_i^0},\qquad
 \bar r_a(t)=\frac{1}{n}\sum_{i=1}^{n}r_{ia}(t).
 \label{eq:seed-reduction}
\end{equation}
The paired advantage of generated program $m$ over baseline $b$ is
\begin{equation}
 \Delta_{m,b}(t)=\frac{1}{n}\sum_{i=1}^{n}[r_{im}(t)-r_{ib}(t)].
 \label{eq:paired-advantage}
\end{equation}
Positive values in~\eqref{eq:paired-advantage} favor $m$ and are measured in percentage points. For example, reductions of 8\% and 6\% give an advantage of $+2$ percentage points. 

Methods share each instance's starting solution and solver seeds. Each receives the same routing budget after the common starting solution has been constructed. 
To quantify instance variability, we resample test instances with replacement within each panel and recompute the mean paired advantage for 20,000 resamples. 
Appendix~\ref{app:metrics} gives the sampling procedure and interval construction.

\subsection{Discovery and validation}

Our automatic heuristic design (AHD) process has two phases, discovery and validation.

During discovery, an LLM proposes a short design hypothesis and a complete program at each iteration.
The language model receives fixed task instructions and a \emph{discovery history} of earlier hypotheses, validity outcomes, measured reductions and decisions about which programs to retain for comparison.
This history differs from the within-run routing history $R$ above. We withhold the source code of earlier proposals. A failed proposal may receive at most one correction.

We start with the first generated program $h_0$ as the program retained for comparison, $h_\star$.
For each valid proposal $h$, we use a 60-second screen per instance on the discovery panel to obtain its mean reduction $r_{60}(h)$.
The proposal advances if its screening score is at most $\varepsilon$ percentage points below that of $h_\star$.

For proposals that advance, we run a fresh evaluation with 120 seconds per instance on the same panel from the same starting routes. This gives the mean reduction $r_{120}(h)$. We replace $h_\star$ if the new program's 120-second score exceeds the retained program's by at least $\delta$ percentage points. We record each proposal's outcome and repeat within the discovery budget.

After discovery, we shortlist twelve programs with distinct source code based on their 120-second scores. We then reevaluate these programs on a separate validation panel.
We rank programs by mean reduction and break exact ties in favor of earlier proposals. Algorithm~\ref{alg:autoresearch} uses $G$ for the language-model generator and $H$ for the discovery history.
The discovery budget sets the proposal cap and limits the time, compute, and API spending available for discovery. Appendix~\ref{app:evaluation-protocol} gives the evaluation settings and shortlist rules.

\begin{algorithm}[!htbp]
\SetAlgoVlined
\DontPrintSemicolon
\KwIn{Initial program $h_0$, generator $G$, discovery and validation panels, discovery budget, margins $\varepsilon,\delta$}
\textbf{Discovery}\;
Evaluate $h_0$ at both discovery stages and initialize $H$\;
$h_\star\leftarrow h_0$\;
\While{the discovery budget allows another iteration}{
    Generate a hypothesis and program $h$ using $G(H)$\;
    \If{$h$ passes code checks and returns valid delegations on sample states}{
        $r_{60}(h)\leftarrow$ mean reduction from the 60-second screen\;
        \If{$r_{60}(h)\ge r_{60}(h_\star)-\varepsilon$}{
            $r_{120}(h)\leftarrow$ mean reduction from a fresh 120-second full evaluation\;
            \If{$r_{120}(h)\ge r_{120}(h_\star)+\delta$}{
                $h_\star\leftarrow h$\;
            }
        }
    }
    Record the proposal and its outcomes in $H$\;
}
\textbf{Validation}\;
Shortlist 12 distinct program sources by $r_{120}$, including the source of $h_\star$\;
Evaluate the shortlist on the separate validation panel\;
\KwOut{Program with the highest mean validation reduction}
\caption{Program discovery and validation}
\label{alg:autoresearch}
\end{algorithm}

The selected program is then tested without source changes on held-out instances. Section~\ref{sec:experiments} reports the discovered and selected policies. 
Appendix~\ref{app:heuristics} gives their original hypotheses and rule summaries.

\newpage
\section{Experimental Setup and Results}
\label{sec:experiments}

\subsection{Instances and baselines}
\label{sec:test-panels}

We use two families of synthetic CVRP instances. \textbf{CVRP-S} follows the instance distributions used by L2D and supplies the discovery, validation, and in-distribution test sets. \textbf{CVRP-X} varies additional features of the routing problem and serves as the out-of-distribution test family.

\paragraph{CVRP-S.}
All instances have 500 customers. Uniform instances sample customer locations across the unit square, clustered instances concentrate customers around several centers, and mixed instances combine both patterns. The clustered and mixed instances are divided equally among settings with three, five, or seven clusters. Across all three customer distributions, we draw each customer's demand uniformly from the integers 1 through 9 and set vehicle capacity to 50. Discovery and validation each use a separate set of 90 instances, with 30 per customer distribution. The test set contains 270 further instances, with 90 per customer distribution.

\paragraph{CVRP-X.}
We generate instances using the \href{https://galgos.inf.puc-rio.br/cvrplib/en/cvrp/xml100}{CVRPLIB generator} for the X-family design of~\citet{uchoa2017benchmark}. The generator uses random, clustered, and mixed customer layouts on a grid. It also varies depot placement, customer demand, and vehicle capacity. We generate 90 instances at each of three sizes, with $N=500$, $1000$, and $2000$ customers. Appendix~\ref{app:protocol} describes the starting solutions, repair settings, and distance convention for CVRP-X.
Customer-location generation and example layouts for both families appear in Appendix~\ref{app:instance-design}.

Discovery and validation use LKH repairs on 500-customer CVRP-S instances. We test CVRP-S with LKH and CVRP-X with both LKH and HGS, using one repair solver per run. 
CVRP-S tests receive 120 seconds; CVRP-X tests receive 300, 600 and 1200 seconds for $N=500,1000,2000$, respectively. These are elapsed routing budgets measured from the common supplied starting solution. On each instance, all methods use the same saved starting solution and the same three solver seeds.

\paragraph{Baselines.}
We compare the generated programs with three baselines from L2D~\citep{li2021learning}. The baseline labeled \emph{Centroid} uses the centroid-min-count heuristic. It groups nearby routes around a center route and gives priority to groups whose center-route customers have been selected less often. The neural baselines use released checkpoints, which are saved trained models. \LTwoDUniform{} was trained on uniform instances, and \LTwoDClusteredMixed{} was trained on clustered and mixed instances.
For each repair solver, we use the corresponding pretrained L2D checkpoints at every test size. The evaluated generated programs retain their source code when tested with HGS. The comparison includes each method's candidate construction and overhead. Appendix~\ref{app:baseline-construction} gives the implementation details.

We also run LKH and HGS directly on full instances with longer budgets to obtain secondary solution-quality references. Appendix~\ref{app:reference-evidence} reports those comparisons and their timing conventions. Results from these whole-instance runs are excluded from discovery and validation.

\subsection{Discovery and validation results}

\begin{figure}[t]
 \centering
 \includegraphics[width=\linewidth]{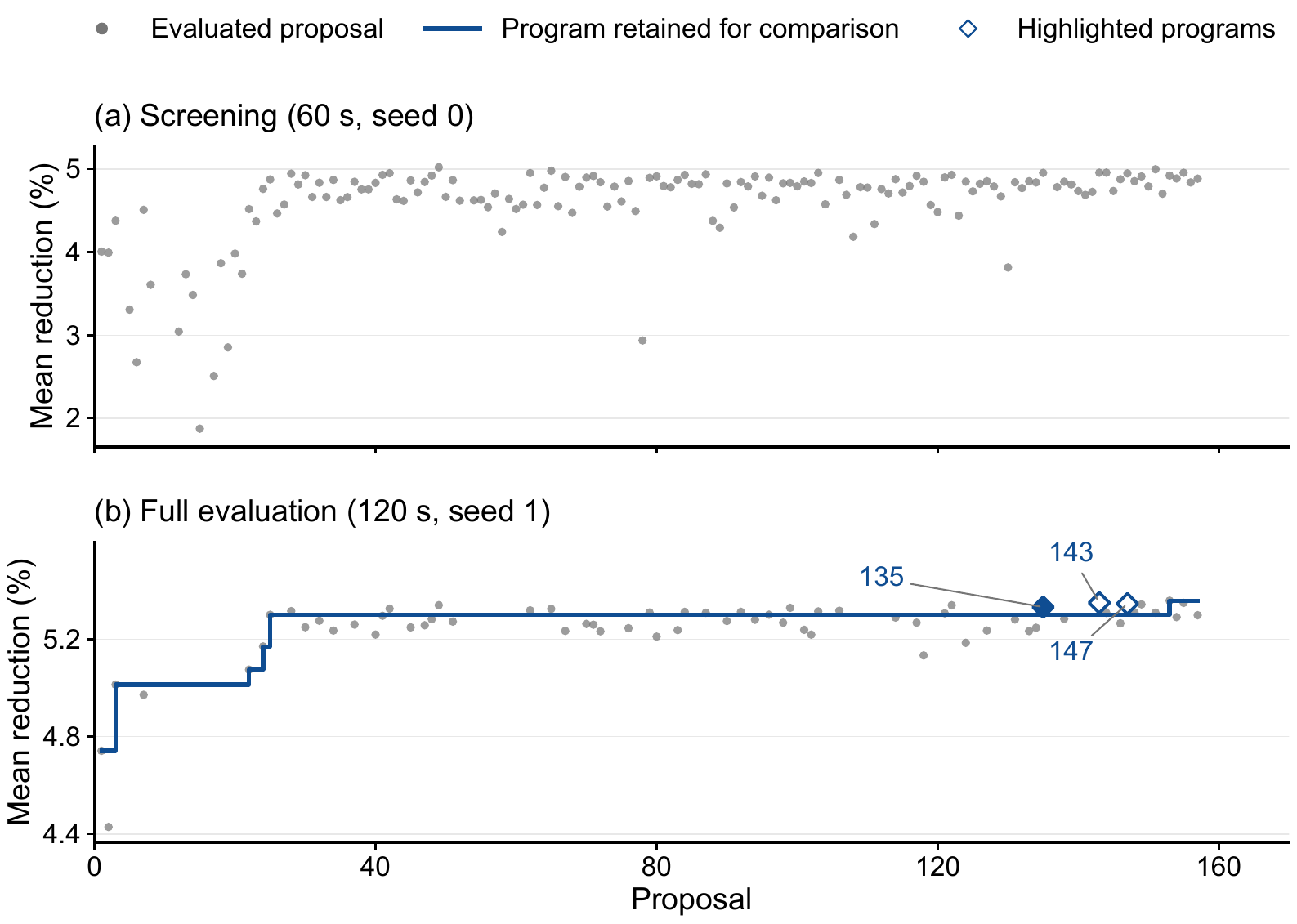}
 \caption{Proposal evaluations on the CVRP-S discovery panel. The step line tracks the full-evaluation reduction of the program retained for comparison. Program numbers are discovery proposal IDs. The filled diamond marks the program selected on validation, and the open diamonds mark the two exploratory programs.}
 \label{fig:discovery}
\end{figure}

The discovery campaign made 157 proposals, including unsuccessful attempts, out of a 200-proposal allowance. It completed 149 screens and 65 full 120-second evaluations. GPT-6 Astra with high reasoning effort generated the programs.
The highest mean reduction in the 120-second discovery evaluations was 5.36\%, compared with 4.74\% for the initial program (Figure~\ref{fig:discovery}).

We identify each generated program by its discovery proposal number. Table~\ref{tab:validation-top-three} shows the three highest-ranked programs on the separate validation panel. We selected policy 135, which achieved the highest mean reduction, as the primary program for testing. Section~\ref{sec:validation-test} examines policies 143 and 147 as exploratory programs. Table~\ref{tab:validation} in Appendix~\ref{app:supporting-results} gives the full ranking of all twelve programs.

\begin{table}[!htbp]
 \centering
 \small
 \caption{Mean reduction for the three highest-ranked programs on validation.}
 \label{tab:validation-top-three}
 \begin{tabular}{rrrl}
  \hline
  Validation rank & Policy & Mean reduction (\%) & Role \\
  \hline
  1 & 135 & \textbf{5.24} & Primary \\
  2 & 143 & 5.20 & Exploratory \\
  3 & 147 & 5.18 & Exploratory \\
  \hline
 \end{tabular}
\end{table}

\subsection{Cost reduction and transfer results}
\label{sec:transfer-results}

Table~\ref{tab:primary} reports policy 135's paired advantages in percentage points for three baselines in each of seven test panels. Table~\ref{tab:reduction-all} in Appendix~\ref{app:objective-choice} gives each method's mean reduction.

On CVRP-S, policy 135 achieved cost reductions comparable in magnitude to those of the L2D models. Its mean reduction was 0.07--0.17 percentage points below the three baselines, and all three adjusted intervals lay below zero.

The clearest evidence of transfer comes from CVRP-X with LKH. Policy 135 achieved greater mean reduction than both L2D models at all three sizes. The advantages ranged from 0.24 to 0.62 percentage points, with all six adjusted intervals above zero. The advantage was already present at 500 customers, the size used during discovery.

\begin{figure}[t]
 \centering
 \includegraphics[width=\linewidth]{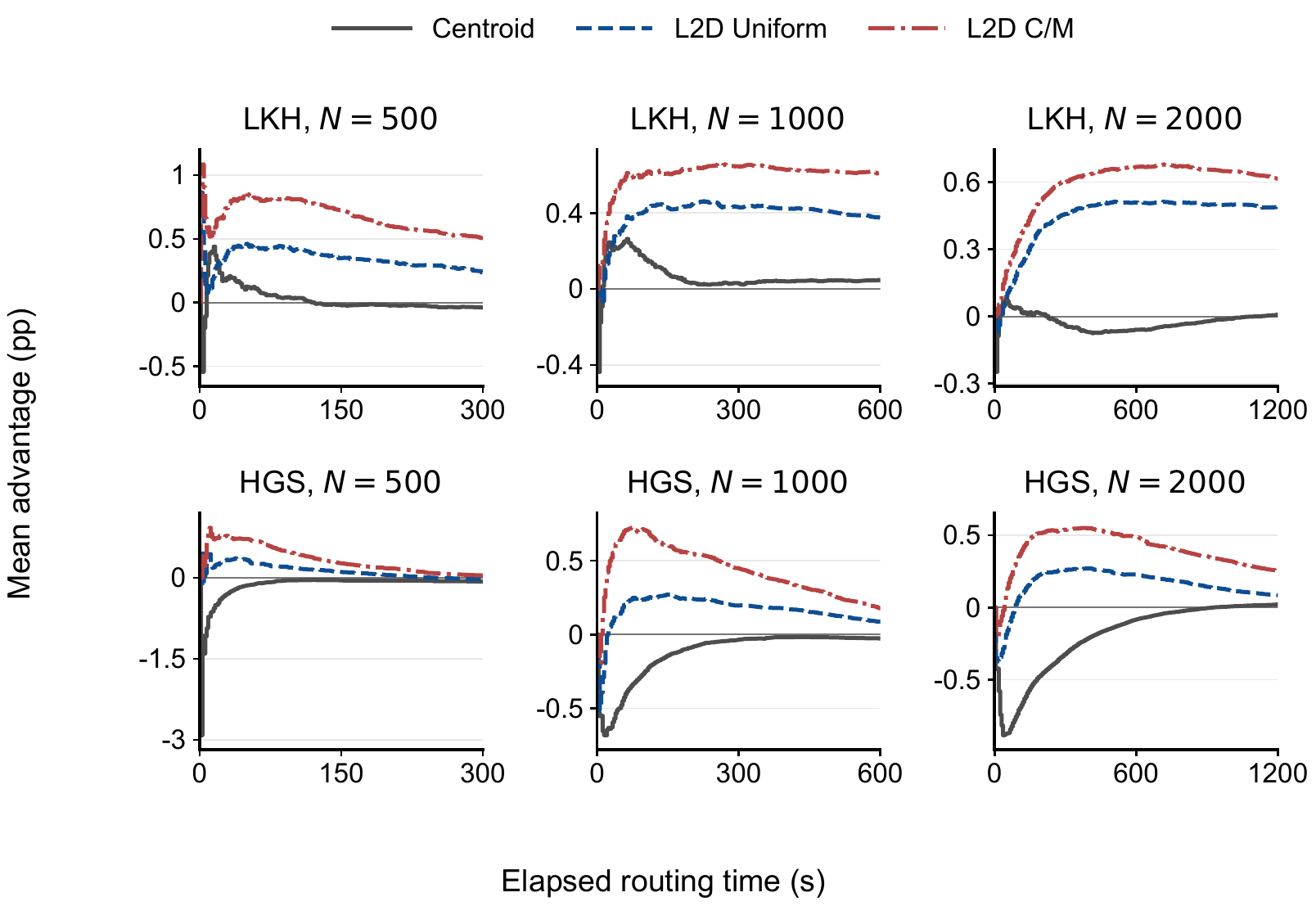}
 \caption{Policy 135's mean paired advantage over each baseline on CVRP-X; positive values favor 135. Means only; panel y-scales differ. Final-budget adjusted intervals appear in Table~\ref{tab:primary-ci}.}
 \label{fig:primary-advantage-trajectories}
\end{figure}

The program also showed an advantage when both the instance family and the repair solver changed. 
With HGS, policy 135 improved on \LTwoDClusteredMixed{} by 0.18 and 0.25 percentage points at 1000 and 2000 customers, respectively, with both adjusted intervals above zero. The interval at 500 customers and all intervals against \LTwoDUniform{} included zero. All six CVRP-X intervals against Centroid also included zero. Figure~\ref{fig:primary-advantage-trajectories} shows how the paired advantages evolve within each panel's routing budget.

\begin{table}[t]
\centering
\caption{Policy 135: paired advantages (percentage points).}
\label{tab:primary}
\label{tab:primary-ci}
\small
\setlength{\tabcolsep}{3pt}
\renewcommand{\arraystretch}{1.07}
\begin{tabular}{@{}lccc@{}}
\hline
Panel & Centroid & \LTwoDUniform{} & \LTwoDClusteredMixed{} \\
 & \multicolumn{3}{c}{Mean [95\% simultaneous interval]} \\
\hline
S500/LKH & \(-0.07\,[-0.12,-0.02]\) & \(-0.15\,[-0.20,-0.09]\) & \(-0.17\,[-0.22,-0.12]\) \\
\noalign{\vskip3pt}
X500/LKH & \(-0.04\,[-0.18,+0.10]\) & \(\boldsymbol{+0.24}\,\boldsymbol{[+0.07,+0.43]}\) & \(\boldsymbol{+0.51}\,\boldsymbol{[+0.24,+0.84]}\) \\
X1000/LKH & \(\boldsymbol{+0.05}\,[-0.14,+0.22]\) & \(\boldsymbol{+0.38}\,\boldsymbol{[+0.21,+0.56]}\) & \(\boldsymbol{+0.61}\,\boldsymbol{[+0.41,+0.80]}\) \\
X2000/LKH & \(\boldsymbol{+0.01}\,[-0.10,+0.11]\) & \(\boldsymbol{+0.49}\,\boldsymbol{[+0.33,+0.66]}\) & \(\boldsymbol{+0.62}\,\boldsymbol{[+0.43,+0.83]}\) \\
\noalign{\vskip3pt}
X500/HGS & \(-0.07\,[-0.26,+0.11]\) & \(-0.03\,[-0.18,+0.11]\) & \(\boldsymbol{+0.04}\,[-0.10,+0.19]\) \\
X1000/HGS & \(-0.03\,[-0.18,+0.14]\) & \(\boldsymbol{+0.09}\,[-0.07,+0.26]\) & \(\boldsymbol{+0.18}\,\boldsymbol{[+0.02,+0.36]}\) \\
X2000/HGS & \(\boldsymbol{+0.02}\,[-0.14,+0.18]\) & \(\boldsymbol{+0.08}\,[-0.05,+0.22]\) & \(\boldsymbol{+0.25}\,\boldsymbol{[+0.11,+0.41]}\) \\
\hline
\end{tabular}
\par\smallskip
\begin{minipage}{0.98\linewidth}\footnotesize X1000/HGS denotes 1,000-customer CVRP-X instances evaluated with HGS. Positive values favor policy 135. Bold marks positive means and intervals entirely above zero. Bonferroni-adjusted intervals target nominal 95\% simultaneous coverage across 21 comparisons.\end{minipage}
\end{table}

\subsection{Exploratory programs}
\label{sec:validation-test}

We also examine whether discovery produced other programs that transfer well. Policies 143 and 147 ranked second and third on validation. We added them as exploratory comparisons after inspecting the transfer results. Policy 135 remains the primary program.

Both exploratory programs achieved higher observed mean reductions than policy 135 in all seven test panels. The change in ranking also occurred on held-out CVRP-S instances, showing that the ranking from a single validation panel can change even within the same instance family. Table~\ref{tab:selection-transfer} in Appendix~\ref{app:validation-test-means} gives the validation and test means.

The LKH transfer result extended to both exploratory programs. On CVRP-X, both outperformed the two L2D models at every size and Centroid at 1000 customers, with adjusted intervals above zero. With HGS, the adjusted intervals lay above zero in five of policy 143's six comparisons with L2D and in three of policy 147's six comparisons. On CVRP-S, both exploratory programs trailed the L2D models. Table~\ref{tab:exploratory-ci} reports all paired advantages and adjusted intervals, with Bonferroni correction applied separately to each policy.

\begin{table}[!htbp]
\centering
\caption{Exploratory policies 143 and 147: paired advantages (percentage points).}
\label{tab:exploratory-ci}
\small
\setlength{\tabcolsep}{3pt}
\renewcommand{\arraystretch}{1.07}
\begin{tabular}{@{}lccc@{}}
\hline
Panel & Centroid & \LTwoDUniform{} & \LTwoDClusteredMixed{} \\
 & \multicolumn{3}{c}{Mean [95\% simultaneous interval]} \\
\hline
\multicolumn{4}{l}{\textbf{Policy 143 (rank 2)}} \\[2pt]
S500/LKH & \(-0.01\,[-0.06,+0.03]\) & \(-0.09\,[-0.15,-0.03]\) & \(-0.11\,[-0.16,-0.06]\) \\
\noalign{\vskip3pt}
X500/LKH & \(\boldsymbol{+0.09}\,[-0.06,+0.24]\) & \(\boldsymbol{+0.37}\,\boldsymbol{[+0.22,+0.54]}\) & \(\boldsymbol{+0.64}\,\boldsymbol{[+0.38,+0.95]}\) \\
X1000/LKH & \(\boldsymbol{+0.19}\,\boldsymbol{[+0.06,+0.34]}\) & \(\boldsymbol{+0.52}\,\boldsymbol{[+0.36,+0.70]}\) & \(\boldsymbol{+0.75}\,\boldsymbol{[+0.55,+0.95]}\) \\
X2000/LKH & \(\boldsymbol{+0.05}\,[-0.05,+0.16]\) & \(\boldsymbol{+0.53}\,\boldsymbol{[+0.38,+0.70]}\) & \(\boldsymbol{+0.66}\,\boldsymbol{[+0.47,+0.86]}\) \\
\noalign{\vskip3pt}
X500/HGS & \(\boldsymbol{+0.02}\,[-0.15,+0.21]\) & \(\boldsymbol{+0.06}\,[-0.08,+0.22]\) & \(\boldsymbol{+0.14}\,\boldsymbol{[+0.0046,+0.28]}\) \\
X1000/HGS & \(\boldsymbol{+0.10}\,[-0.05,+0.26]\) & \(\boldsymbol{+0.21}\,\boldsymbol{[+0.06,+0.38]}\) & \(\boldsymbol{+0.30}\,\boldsymbol{[+0.15,+0.48]}\) \\
X2000/HGS & \(\boldsymbol{+0.09}\,[-0.07,+0.26]\) & \(\boldsymbol{+0.15}\,\boldsymbol{[+0.02,+0.30]}\) & \(\boldsymbol{+0.33}\,\boldsymbol{[+0.18,+0.48]}\) \\
\noalign{\vskip5pt}
\hline
\multicolumn{4}{l}{\textbf{Policy 147 (rank 3)}} \\[2pt]
S500/LKH & \(-0.04\,[-0.09,+0.0047]\) & \(-0.12\,[-0.18,-0.05]\) & \(-0.14\,[-0.19,-0.10]\) \\
\noalign{\vskip3pt}
X500/LKH & \(\boldsymbol{+0.07}\,[-0.08,+0.20]\) & \(\boldsymbol{+0.35}\,\boldsymbol{[+0.20,+0.53]}\) & \(\boldsymbol{+0.62}\,\boldsymbol{[+0.35,+0.95]}\) \\
X1000/LKH & \(\boldsymbol{+0.18}\,\boldsymbol{[+0.05,+0.33]}\) & \(\boldsymbol{+0.51}\,\boldsymbol{[+0.35,+0.71]}\) & \(\boldsymbol{+0.74}\,\boldsymbol{[+0.54,+0.95]}\) \\
X2000/LKH & \(\boldsymbol{+0.06}\,[-0.03,+0.16]\) & \(\boldsymbol{+0.54}\,\boldsymbol{[+0.39,+0.71]}\) & \(\boldsymbol{+0.67}\,\boldsymbol{[+0.49,+0.88]}\) \\
\noalign{\vskip3pt}
X500/HGS & \(-0.03\,[-0.22,+0.18]\) & \(\boldsymbol{+0.0041}\,[-0.14,+0.19]\) & \(\boldsymbol{+0.08}\,[-0.08,+0.26]\) \\
X1000/HGS & \(\boldsymbol{+0.05}\,[-0.11,+0.22]\) & \(\boldsymbol{+0.16}\,\boldsymbol{[+0.0021,+0.34]}\) & \(\boldsymbol{+0.25}\,\boldsymbol{[+0.10,+0.43]}\) \\
X2000/HGS & \(\boldsymbol{+0.06}\,[-0.09,+0.22]\) & \(\boldsymbol{+0.12}\,[-0.01,+0.26]\) & \(\boldsymbol{+0.29}\,\boldsymbol{[+0.15,+0.44]}\) \\
\hline
\end{tabular}
\par\smallskip
\begin{minipage}{0.98\linewidth}\footnotesize Positive values favor the named policy. Bold marks positive means and intervals entirely above zero. Bonferroni-adjusted intervals target nominal 95\% simultaneous coverage across each policy's 21 comparisons. Ranks are from validation. Nonzero values that would round to zero are shown to four decimals.\end{minipage}
\end{table}

\subsection{Computing resources and cost}
\label{sec:resources}

Routing evaluations ran on AMD EPYC 9654 CPUs. Each worker used one physical core and ran the solver single-threaded. Discovery and transfer used up to 90 workers. Appendix~\ref{app:protocol} reports the memory allocations and solver settings.

The campaign incurred \$109.79 in accounted charges, with another \$16.57 reserved and unsettled. Discovery allocations totaled 22~h 21~min 02~s, excluding queueing (Table~\ref{tab:discovery}). Once a program is selected, routing with it requires no further LLM calls.

\section{Discussion}
\label{sec:discussion}

The results show that delegation programs discovered on 500-customer CVRP-S instances with LKH achieve higher mean reduction than the evaluated L2D models in most transfer comparisons, without source changes across the tested sizes and repair solvers.

On held-out CVRP-S instances, the primary policy does not demonstrate an advantage over the three baselines. Across the six CVRP-X panels, however, policy 135 achieves greater mean reduction than \LTwoDUniform{} in five panels, with Bonferroni-adjusted intervals entirely above zero in three.
Against \LTwoDClusteredMixed{}, it achieves greater mean reduction in all six panels, with adjusted intervals entirely above zero in five.

Exploratory programs 143 and 147, ranked second and third on validation, achieve higher observed mean reduction than policy 135 on all seven test panels (Section~\ref{sec:validation-test}).
Additionally, these programs achieve greater mean reduction than both L2D models in all six CVRP-X panels. Policy 143 also exceeds Centroid in all six panels, and policy 147 does so in five.

Our validation panel uses only 500-customer CVRP-S instances with LKH. A future campaign could compare the shortlisted programs on separate validation panels covering different instance families, sizes and repair solvers expected in use before selecting one program for testing.

Beyond vehicle routing, LLM-generated algorithms have been studied in inventory control, queueing control and assortment optimization~\citep{baek2026llms}. In warehouse multi-agent pathfinding (MAPF), large neighborhood search replans paths for selected groups of agents while keeping other paths fixed~\citep{li2021mapflns}.
A program generated through AHD could receive the current paths and choose which agents to replan, while a fixed planner constructs the replacement paths. Applying the procedure would require a new evaluation pipeline that checks collisions and compares path costs within a fixed planning budget.
\section{Conclusion}
\label{sec:conclusion}

We extend L2D's select-solve-merge framework~\citep{li2021learning} with hypothesis-driven language-model search for reusable delegation programs. The discovered programs transfer without source changes to generated CVRP-X instances, which are not used during L2D training or AHD. The tests cover changes in instance distributions, sizes and repair solvers.

Once a program has been produced during discovery, the routing procedure can run without further LLM assistance. The original design hypotheses provide a record of the proposed rules for inspection and reuse (Appendix~\ref{app:heuristics}). Further experiments should repeat discovery campaigns and test reuse of the AHD procedure in other supply-chain problems.

\bibliographystyle{plainnat}
\bibliography{references}

@inproceedings{li2021learning,
  title={Learning to Delegate for Large-Scale Vehicle Routing},
  author={Li, Sirui and Yan, Zhongxia and Wu, Cathy},
  booktitle={Advances in Neural Information Processing Systems},
  volume={34}, pages={26198--26211},
  year={2021},
  url={https://papers.nips.cc/paper/2021/hash/dc9fa5f217a1e57b8a6adeb065560b38-Abstract.html}
}

@inproceedings{shaw1998using,
  title={Using Constraint Programming and Local Search Methods to Solve Vehicle Routing Problems},
  author={Shaw, Paul},
  booktitle={Principles and Practice of Constraint Programming---CP98},
  series={Lecture Notes in Computer Science}, volume={1520}, pages={417--431}, year={1998},
  doi={10.1007/3-540-49481-2_30},
  url={https://link.springer.com/chapter/10.1007/3-540-49481-2_30}
}

@incollection{taillard2002popmusic,
  title={{POPMUSIC}---Partial Optimization Metaheuristic under Special Intensification Conditions},
  author={Taillard, {\'E}ric D. and Voss, Stefan},
  booktitle={Essays and Surveys in Metaheuristics},
  series={Operations Research/Computer Science Interfaces Series}, volume={15}, pages={613--629},
  publisher={Springer}, year={2002},
  doi={10.1007/978-1-4615-1507-4_27},
  url={https://link.springer.com/chapter/10.1007/978-1-4615-1507-4_27}
}

@article{santini2023decomposition,
  title={Decomposition Strategies for Vehicle Routing Heuristics},
  author={Santini, Alberto and Schneider, Michael and Vidal, Thibaut and Vigo, Daniele},
  journal={INFORMS Journal on Computing}, volume={35}, number={3}, pages={543--559}, year={2023},
  doi={10.1287/ijoc.2023.1288}, url={https://doi.org/10.1287/ijoc.2023.1288}
}

@inproceedings{kool2019attention,
  title={Attention, Learn to Solve Routing Problems!},
  author={Kool, Wouter and van Hoof, Herke and Welling, Max},
  booktitle={International Conference on Learning Representations}, year={2019},
  url={https://openreview.net/forum?id=ByxBFsRqYm}
}

@inproceedings{hottung2020neural,
  title={Neural Large Neighborhood Search for the Capacitated Vehicle Routing Problem},
  author={Hottung, Andr{\'e} and Tierney, Kevin},
  booktitle={ECAI 2020}, pages={443--450}, year={2020},
  doi={10.3233/FAIA200124}, url={https://doi.org/10.3233/FAIA200124}
}

@inproceedings{ye2024glop,
  title={{GLOP}: Learning Global Partition and Local Construction for Solving Large-Scale Routing Problems in Real-Time},
  author={Ye, Haoran and Wang, Jiarui and Liang, Helan and Cao, Zhiguang and Li, Yong and Li, Fanzhang},
  booktitle={Proceedings of the AAAI Conference on Artificial Intelligence},
  volume={38}, pages={20284--20292}, year={2024},
  doi={10.1609/aaai.v38i18.30009}, url={https://ojs.aaai.org/index.php/AAAI/article/view/30009}
}

@inproceedings{bi2022amdkd,
  title={Learning Generalizable Models for Vehicle Routing Problems via Knowledge Distillation},
  author={Bi, Jieyi and Ma, Yining and Wang, Jiahai and Cao, Zhiguang and Chen, Jinbiao and Sun, Yuan and Chee, Yeow Meng},
  booktitle={Advances in Neural Information Processing Systems},
  volume={35}, pages={31226--31238}, year={2022},
  url={https://papers.nips.cc/paper_files/paper/2022/hash/ca70528fb11dc8086c6a623da9f3fee6-Abstract-Conference.html}
}

@inproceedings{yu2026coeks,
  title={{Combination-of-Experts} with Knowledge Sharing for Cross-Task Vehicle Routing Problems},
  author={Yu, Zikang and Chen, Jinbiao and Wang, Jiahai},
  booktitle={International Conference on Learning Representations}, year={2026},
  url={https://proceedings.iclr.cc/paper_files/paper/2026/hash/4572bc2f514e627914cbe60d0398a2d1-Abstract-Conference.html}
}

@article{romeraparedes2024funsearch,
  title={Mathematical discoveries from program search with large language models},
  author={Romera-Paredes, Bernardino and Barekatain, Mohammadamin and Novikov, Alexander and Balog, Matej and Kumar, M. Pawan and Dupont, Emilien and Ruiz, Francisco J. R. and Ellenberg, Jordan S. and Wang, Pengming and Fawzi, Omar and Kohli, Pushmeet and Fawzi, Alhussein},
  journal={Nature}, volume={625}, pages={468--475}, year={2024},
  doi={10.1038/s41586-023-06924-6},
  url={https://doi.org/10.1038/s41586-023-06924-6}
}

@inproceedings{liu2024eoh,
  title={Evolution of Heuristics: Towards Efficient Automatic Algorithm Design Using Large Language Model},
  author={Liu, Fei and Tong, Xialiang and Yuan, Mingxuan and Lin, Xi and Luo, Fu and Wang, Zhenkun and Lu, Zhichao and Zhang, Qingfu},
  booktitle={Proceedings of the 41st International Conference on Machine Learning},
  series={Proceedings of Machine Learning Research}, volume={235}, pages={32201--32223}, year={2024},
  url={https://proceedings.mlr.press/v235/liu24bs.html}
}

@inproceedings{ye2024reevo,
  title={{ReEvo}: Large Language Models as Hyper-Heuristics with Reflective Evolution},
  author={Ye, Haoran and Wang, Jiarui and Cao, Zhiguang and Berto, Federico and Hua, Chuanbo and Kim, Haeyeon and Park, Jinkyoo and Song, Guojie},
  booktitle={Advances in Neural Information Processing Systems}, volume={37}, pages={43571--43608}, year={2024},
  doi={10.52202/079017-1381},
  url={https://proceedings.neurips.cc/paper_files/paper/2024/hash/4ced59d480e07d290b6f29fc8798f195-Abstract-Conference.html}
}

@article{hottung2026vrpagent,
  title={{VRPAgent}: {LLM}-Driven Discovery of Heuristic Operators for Vehicle Routing Problems},
  author={Hottung, Andr{\'e} and Berto, Federico and Hua, Chuanbo and Gast Zepeda, Nayeli and Wetzel, Daniel and R{\"o}mer, Michael and Ye, Haoran and Zago, Davide and Poli, Michael and Massaroli, Stefano and Park, Jinkyoo and Tierney, Kevin},
  journal={Transactions on Machine Learning Research}, year={2026},
  url={https://arxiv.org/abs/2510.07073v2}
}

@misc{li2025ars,
  title={{ARS}: Automatic Routing Solver with Large Language Models},
  author={Li, Kai and Liu, Fei and Wang, Zhenkun and Tong, Xialiang and Han, Xiongwei and Yuan, Mingxuan and Zhang, Qingfu},
  year={2025}, howpublished={arXiv preprint arXiv:2502.15359},
  url={https://arxiv.org/abs/2502.15359v3}
}

@misc{lai2026specahd,
  title={{SpecAHD}: Localize to Specialize for Automated Heuristic Design in Large-Scale Routing Problems},
  author={Lai, Kezhao and Lai, Yutao and Liu, Hai-Lin},
  year={2026}, howpublished={arXiv preprint arXiv:2607.23676},
  url={https://arxiv.org/abs/2607.23676v1}
}

@misc{xie2026ailsahd,
  title={Enhancing {CVRP} Solver through {LLM}-Driven Automatic Heuristic Design},
  author={Xie, Zhuoliang and Liu, Fei and Wang, Zhenkun and Zhang, Qingfu},
  year={2026}, howpublished={arXiv preprint arXiv:2602.23092},
  doi={10.48550/arXiv.2602.23092}, url={https://arxiv.org/abs/2602.23092v2}
}

@inproceedings{malik2026pyvrpplus,
  title={{PyVRP+}: {LLM}-Driven Metacognitive Heuristic Evolution for Hybrid Genetic Search in Vehicle Routing Problems},
  author={Malik, Manuj and Zhou, Jianan and Chirra, Shashank Reddy and Cao, Zhiguang},
  booktitle={Proceedings of the 25th International Conference on Autonomous Agents and Multiagent Systems},
  pages={2800--2808}, year={2026},
  doi={10.65109/ISFV2587}, url={https://ink.library.smu.edu.sg/sis_research/11392/}
}

@inproceedings{li2021mapflns,
  author = {Jiaoyang Li and Zhe Chen and Daniel Harabor and Peter J. Stuckey and Sven Koenig},
  title = {Anytime Multi-Agent Path Finding via Large Neighborhood Search},
  booktitle = {Proceedings of the Thirtieth International Joint Conference on Artificial Intelligence},
  pages = {4127--4135},
  year = {2021},
  doi = {10.24963/ijcai.2021/568},
  url = {https://ijcai.org/proceedings/2021/568}
}

@techreport{helsgaun2017lkh3,
  author={Helsgaun, Keld},
  title={An Extension of the {Lin--Kernighan--Helsgaun} {TSP} Solver for Constrained Traveling Salesman and Vehicle Routing Problems},
  institution={Roskilde University}, year={2017},
  url={https://forskning.ruc.dk/en/publications/an-extension-of-the-lin-kernighan-helsgaun-tsp-solver-for-constra/}
}

@article{vidal2012hybrid,
  author={Vidal, Thibaut and Crainic, Teodor Gabriel and Gendreau, Michel and Lahrichi, Nadia and Rei, Walter},
  title={A Hybrid Genetic Algorithm for Multidepot and Periodic Vehicle Routing Problems},
  journal={Operations Research}, volume={60}, number={3}, pages={611--624}, year={2012},
  doi={10.1287/opre.1120.1048}, url={https://doi.org/10.1287/opre.1120.1048}
}

@article{vidal2022hgs,
  author={Vidal, Thibaut},
  title={Hybrid genetic search for the {CVRP}: Open-source implementation and {SWAP*} neighborhood},
  journal={Computers \& Operations Research}, volume={140}, pages={105643}, year={2022},
  doi={10.1016/j.cor.2021.105643}, url={https://doi.org/10.1016/j.cor.2021.105643}
}

@inproceedings{sun2025cobench,
  author={Sun, Weiwei and Feng, Shengyu and Li, Shanda and Yang, Yiming},
  title={{CO-Bench}: Benchmarking Language Model Agents in Algorithm Search for Combinatorial Optimization},
  booktitle={Proceedings of the AAAI Conference on Artificial Intelligence},
  volume={40}, pages={33126--33134}, year={2026},
  doi={10.1609/aaai.v40i39.40596},
  url={https://ojs.aaai.org/index.php/AAAI/article/view/40596}
}

@article{gong2026lace,
  author={Gong, Huatian and Wang, Shuaian and Song, Dongping and Sheu, Jiuh-Biing and Yan, Ran},
  title={Large language models discover complementary heuristics for combinatorial optimization},
  journal={Nature Machine Intelligence}, volume={8}, pages={1569--1580}, year={2026},
  doi={10.1038/s42256-026-01307-8}, url={https://doi.org/10.1038/s42256-026-01307-8}
}

@inproceedings{liu2026mths,
  author={Liu, Fei and Zhang, Rui and Yao, Shunyu and Hu, Qinglong and Zheng, Kefeng and Lu, Zhichao and Zhang, Qingfu},
  title={Hierarchical Representations for Cross-task Automated Heuristic Design using {LLMs}},
  booktitle={Proceedings of the 43rd International Conference on Machine Learning},
  series={Proceedings of Machine Learning Research}, volume={306}, pages={79907--79941}, year={2026},
  url={https://proceedings.mlr.press/v306/liu26ge.html}
}

@misc{baek2026llms,
  author={Baek, Jackie},
  title={{LLMs} Can Design Near-Optimal {OR} Algorithms},
  year={2026}, howpublished={arXiv preprint arXiv:2608.27296},
  doi={10.48550/arXiv.2608.27296},
  url={https://arxiv.org/abs/2608.27296v3}
}

@article{uchoa2017benchmark,
  author={Uchoa, Eduardo and Pecin, Diego and Pessoa, Artur and Poggi, Marcus and Vidal, Thibaut and Subramanian, Anand},
  title={New benchmark instances for the Capacitated Vehicle Routing Problem},
  journal={European Journal of Operational Research},
  volume={257}, number={3}, pages={845--858}, year={2017},
  doi={10.1016/j.ejor.2016.08.012},
  url={https://doi.org/10.1016/j.ejor.2016.08.012}
}

\appendix
\section{Objective choice}
\label{app:objective-choice}
\label{app:raw-costs}

Mean start-normalized cost reduction compares percentage reductions on each instance, regardless of the instance's cost scale. We use this metric to compare a delegation program's relative gains across instances. Maximizing reduction is equivalent to minimizing mean final cost normalized by each instance's starting cost. The comparison here is with \emph{raw mean final cost}, which averages distances without this normalization. The same percentage reduction contributes more absolute savings on an instance with a higher starting cost. Within each CVRP-X size panel, the 95th-percentile starting cost is 7.92--10.38 times the 5th-percentile starting cost.

Let $C_i^0$ be instance $i$'s common starting cost, $\bar C_{ia}$ method $a$'s seed-averaged final cost, and $n$ the number of instances. Mean normalized final cost is
\begin{equation}
 F_a^{\rm norm}=\frac1n\sum_{i=1}^n\frac{\bar C_{ia}}{C_i^0}
 =1-\frac{\bar r_a}{100},
 \label{eq:normalized-final-equivalence}
\end{equation}
where $\bar r_a$ is mean percentage reduction from~\eqref{eq:seed-reduction}. Raw (unnormalized) mean final cost is
\begin{equation}
 \mathcal C_a=\frac1n\sum_{i=1}^n\bar C_{ia}.
 \label{eq:mean-final-cost}
\end{equation}

Tables~\ref{tab:reduction-all} and~\ref{tab:raw-final-cost-all} give descriptive means for the same runs, with matching panels and method order. Each averages the three solver seeds within an instance, then gives instances equal weight. Policy 135 has higher mean reduction and lower raw mean final cost than both L2D models in every CVRP-X/LKH panel. Centroid has lower raw mean final cost than policy 135 in all six CVRP-X panels, including three where policy 135 has higher mean reduction. Adjusted reduction intervals appear in Table~\ref{tab:primary-ci}.

\begin{table}[!htbp]
\centering
\small
\setlength{\tabcolsep}{4 pt}
\caption{Mean reduction for all six methods (\%).}
\label{tab:reduction-all}
\begin{tabular}{lrrrrrr}
\hline
Panel & 135 & 143 & 147 & Centroid & \LTwoDUniform{} & \LTwoDClusteredMixed{} \\
\hline
CVRP-S 500/LKH & 5.2014 & 5.2560 & 5.2267 & 5.2676 & 5.3484 & \textbf{5.3666} \\
CVRP-X 500/LKH & 7.5110 & \textbf{7.6397} & 7.6212 & 7.5490 & 7.2735 & 7.0039 \\
CVRP-X 1000/LKH & 4.1048 & \textbf{4.2460} & 4.2373 & 4.0576 & 3.7285 & 3.4968 \\
CVRP-X 2000/LKH & 2.7091 & 2.7530 & \textbf{2.7648} & 2.7021 & 2.2207 & 2.0938 \\
CVRP-X 500/HGS & 7.8440 & \textbf{7.9395} & 7.8802 & 7.9146 & 7.8760 & 7.8036 \\
CVRP-X 1000/HGS & 4.5296 & \textbf{4.6563} & 4.6065 & 4.5562 & 4.4442 & 4.3516 \\
CVRP-X 2000/HGS & 3.1619 & \textbf{3.2351} & 3.1996 & 3.1416 & 3.0803 & 2.9085 \\
\hline
\end{tabular}
\par\smallskip
\begin{minipage}{0.98\linewidth}\footnotesize Higher is better; bold marks row maxima. Policy 135 is primary; 143 and 147 are exploratory.\end{minipage}
\end{table}

\begin{table}[!htbp]
\centering
\small
\setlength{\tabcolsep}{4 pt}
\caption{Raw mean final route costs for all six methods.}
\label{tab:raw-final-cost-all}
\begin{tabular}{lrrrrrr}
\hline
Panel & 135 & 143 & 147 & Centroid & \LTwoDUniform{} & \LTwoDClusteredMixed{} \\
\hline
CVRP-S 500/LKH & 56.8317 & 56.8000 & 56.8152 & 56.7896 & 56.7405 & \textbf{56.7331} \\
CVRP-X 500/LKH & 73.5889 & 73.4878 & 73.4936 & \textbf{73.4494} & 73.7899 & 74.0497 \\
CVRP-X 1000/LKH & 140.8941 & 140.6423 & \textbf{140.6375} & 140.6817 & 141.2484 & 141.6377 \\
CVRP-X 2000/LKH & 280.0833 & 279.9599 & 279.8886 & \textbf{279.7306} & 281.0309 & 281.2842 \\
CVRP-X 500/HGS & 73.3577 & 73.2551 & 73.2808 & \textbf{73.1377} & 73.2247 & 73.3476 \\
CVRP-X 1000/HGS & 140.3110 & 140.1387 & 140.2318 & \textbf{139.9861} & 140.2749 & 140.4223 \\
CVRP-X 2000/HGS & 279.0419 & 279.0013 & 279.0130 & \textbf{278.5658} & 278.9525 & 279.4356 \\
\hline
\end{tabular}
\par\smallskip
\begin{minipage}{0.98\linewidth}\footnotesize Continuous-distance units; no normalization by starting cost. Lower is better; bold marks row minima. Instances, starts and routing budgets match Table~\ref{tab:reduction-all}.\end{minipage}
\end{table}

\FloatBarrier

\section{Additional results}
\label{app:supporting-results}

This appendix reports the full validation ranking, CVRP-S reductions by customer distribution and validation and test means for the three evaluated programs. Tables~\ref{tab:primary} and~\ref{tab:exploratory-ci} report the primary and exploratory paired comparisons, respectively. CVRP-S tests contain 270 instances; each CVRP-X size/solver panel contains 90. Each result averages the three seeds within an instance before giving instances equal weight. Policies 143 and 147 remain exploratory, with no test-based reselection of the primary program.

Table~\ref{tab:reduction-all} in Appendix~\ref{app:objective-choice} reports the overall reduction means for every panel.

\begin{table}[!htbp]
\centering
\small
\caption{Validation ranking of the twelve frozen candidates.}
\label{tab:validation}
\begin{tabular}{lrrrr}
\hline
Rank & Policy & Disc. rank & Disc. (\%) & Val. (\%) \\
\hline
1 & 135 (primary) & 8 & 5.3319 & \textbf{5.2403} \\
2 & 143 & 2 & 5.3498 & 5.2002 \\
3 & 147 & 4 & 5.3455 & 5.1835 \\
4 & 65 & 11 & 5.3244 & 5.1800 \\
5 & 155 & 3 & 5.3491 & 5.1798 \\
6 & 153 & 1 & \textbf{5.3583} & 5.1778 \\
7 & 49 & 6 & 5.3396 & 5.1736 \\
8 & 42 & 10 & 5.3253 & 5.1620 \\
9 & 99 & 9 & 5.3287 & 5.1575 \\
10 & 149 & 5 & 5.3431 & 5.1534 \\
11 & 122 & 7 & 5.3395 & 5.1529 \\
12 & 62 & 12 & 5.3186 & 5.1292 \\
\hline
\end{tabular}
\par\smallskip
\begin{minipage}{0.98\linewidth}\footnotesize Full discovery uses one seed, and validation averages over three seeds. Each uses 120 seconds per run on a separate 90-instance panel. Ranks follow validation reduction. Bold marks column maxima. Disc. refers to Discovery; Val. refers to validation.
\end{minipage}
\end{table}

\begin{table}[!htbp]
\centering
\small
\caption{CVRP-S testing: mean reduction (\%).}
\label{tab:astra}
\begin{tabular}{lrrrr}
\hline
Method & Overall & Uniform & Clustered & Mixed \\
\hline
Policy 135 (primary) & 5.2014 & 4.4392 & 6.0529 & 5.1121 \\
Policy 143 & 5.2560 & 4.4686 & 6.1125 & 5.1870 \\
Policy 147 & 5.2267 & 4.4342 & 6.0859 & 5.1601 \\
Centroid-min-count & 5.2676 & 4.4972 & 6.1022 & 5.2035 \\
\LTwoDUniform{} & 5.3484 & \textbf{4.6234} & 6.0708 & \textbf{5.3511} \\
\LTwoDClusteredMixed{} & \textbf{5.3666} & 4.5893 & \textbf{6.2094} & 5.3012 \\
\hline
\end{tabular}
\par\smallskip
\begin{minipage}{0.98\linewidth}\footnotesize Overall: $n=270$; each customer distribution: $n=90$. Descriptive means after 120 seconds; higher is better. Bold marks column maxima.\end{minipage}
\end{table}

\FloatBarrier

\subsection{Validation and test results}
\label{app:validation-test-means}

\begin{table}[!htbp]
\centering
\small
\caption{Validation and test mean reduction (\%).}
\label{tab:selection-transfer}
\begin{tabular}{lrrr}
\hline
Panel & 135 (rank 1) & 143 (rank 2) & 147 (rank 3) \\
\hline
CVRP-S validation & \textbf{5.2403} & 5.2002 & 5.1835 \\
CVRP-S test & 5.2014 & \textbf{5.2560} & 5.2267 \\
CVRP-X 500/LKH & 7.5110 & \textbf{7.6397} & 7.6212 \\
CVRP-X 1000/LKH & 4.1048 & \textbf{4.2460} & 4.2373 \\
CVRP-X 2000/LKH & 2.7091 & 2.7530 & \textbf{2.7648} \\
CVRP-X 500/HGS & 7.8440 & \textbf{7.9395} & 7.8802 \\
CVRP-X 1000/HGS & 4.5296 & \textbf{4.6563} & 4.6065 \\
CVRP-X 2000/HGS & 3.1619 & \textbf{3.2351} & 3.1996 \\
\hline
\end{tabular}
\par\smallskip
\begin{minipage}{0.98\linewidth}\footnotesize Ranks follow validation reduction; subsequent test means are descriptive. Policy 135 remains primary; 143 and 147 are exploratory. Bold marks row maxima.\end{minipage}
\end{table}

\FloatBarrier

\section{Experimental protocol}
\label{app:metrics}
\label{app:protocol}

\subsection{Solver and LLM settings}
\label{app:evaluation-protocol}

To construct a starting solution, we order customers by their angle around the depot. A customer selected using the initialization seed determines where this circular order begins. We divide the order into ten consecutive groups with nearly equal customer counts; these angular sectors need not have equal angular widths. Each sector is initialized separately with at most 100 LKH trials, and its routes are combined with those from the other sectors.

The original initializer failed to produce feasible starting routes for some CVRP-X sectors. Rounding up total sector demand divided by vehicle capacity gives a lower bound on the number of vehicles. Uneven customer demands can require more vehicles than this bound. We therefore construct capacity-feasible routes within each sector before calling LKH.
Best-fit-decreasing considers customers in decreasing demand order and assigns each to the tightest fitting route, opening another route when necessary. Each customer's demand must fit within one vehicle; this construction imposes no fixed fleet cap. Nearest neighbor then orders customer visits within each depot-to-depot route. Customer IDs break demand and distance ties; route indices break packing ties.

LKH receives these constructed routes as an initial tour and uses their route count. We use its returned routes only if they pass feasibility checks and have strictly lower continuous-distance cost than the construction. If LKH finishes without a feasible improvement, we retain the constructed routes. All delegation methods use the same saved starting routes on each instance, whether subsequent repairs use LKH or HGS.

During routing, LKH repairs are cold: they receive the selected customers without the incumbent routes as an initial tour, and run for at most 500 trials. CVRP-S retains the original LKH interface's vehicle-count rule: total selected demand divided by vehicle capacity, rounded up. This is a lower bound based on aggregate capacity; uneven demands can require more vehicles. On CVRP-X, we instead allow up to the number of selected routes, which already provide a feasible partition of those customers. Unused vehicle slots are permitted, and empty routes are removed from the solver output. All methods using LKH follow the same rule. The generated program sources remain unchanged.

HGS repairs use a nominal one-CPU-second stopping threshold, checked between native iterations. A repair is accepted only if its routes pass feasibility and continuous-cost checks, its total cost strictly decreases, and all checks finish within the remaining elapsed budget.

The generator is GPT-6 Astra with high reasoning effort, a 32,768-token output allowance and no tools. In addition to the discovery history described in Section~\ref{sec:methodology}, feedback contains the latest per-case diagnostics and results by customer distribution and, where applicable, by cluster count. It excludes earlier source code and validation/test outcomes. Table~\ref{tab:discovery} gives campaign totals and costs.

Discovery and transfer jobs were allocated 128~GiB of memory. Later recovery and reference jobs were allocated 364~GiB.

\paragraph{Discovery and validation.}
Before routing evaluation, we check the program's syntax and permitted Python features. We then start the program and check that it returns valid delegation lists on representative routing states (Appendix~\ref{app:program-interface}). These checks share a five-second budget. A failed proposal could receive one correction. No correction was requested for the 157 recorded submissions.

Table~\ref{tab:evaluation-settings} gives the per-instance routing budgets and solver seeds for discovery and validation. The margins $\varepsilon$ and $\delta$ are both 0.05 percentage points. Both discovery runs start from the same saved routes. Only completed evaluations supply scores for the comparisons in Algorithm~\ref{alg:autoresearch}.

To build the validation shortlist, we keep the earliest completed 120-second discovery score for each distinct source file. We rank these programs by mean reduction, with exact ties favoring earlier proposals. We take the top twelve and replace the twelfth entry with $h_\star$ if the retained program's source is absent. Validation selects the program with the highest mean reduction on its separate panel, with exact ties again favoring earlier proposals.

\begin{table}[!htbp]
 \centering
 \caption{Evaluation settings for discovery and validation.}
 \label{tab:evaluation-settings}
 \small
 \begin{tabular}{lrr}
 \hline
 Panel & Routing budget & Solver seeds \\
 \hline
 Discovery & 60~s & 0 \\
 Discovery & 120~s & 1 \\
 Validation & 120~s & $0,1,2$ \\
 \hline
 \end{tabular}
\end{table}

\begin{table}[!htbp]
\centering
\small
\caption{Discovery campaign.}
\label{tab:discovery}
\begin{tabular}{lr}
\hline
Quantity & Value \\
\hline
Proposals / allowance & 157 / 200 \\
Completed discovery screens (60 s) & 149 \\
Completed discovery evaluations (120 s) & 65 \\
Initial discovery reduction & 4.7417\% \\
Highest discovery reduction & 5.3583\% \\
Accounted API charges & \$109.79 \\
Unresolved reservations & \$16.57 \\
Charges plus reservations & \$126.36 \\
Discovery allocation elapsed & 22h 21m 02s \\
\hline
\end{tabular}
\par\smallskip
\begin{minipage}{0.98\linewidth}\footnotesize Discovery stages follow Section~\ref{sec:methodology}. API amounts are USD; reservations are unsettled. Allocation time excludes queueing.\end{minipage}
\end{table}

\subsection{Subproblem construction and selection}
\label{app:baseline-construction}

L2D separates candidate construction from selection. A fixed spatial rule forms a candidate delegation around each route centroid by grouping nearby current routes. The trained L2D model predicts the cost after repairing each candidate and ranks candidates by predicted cost reduction.

Centroid-min-count uses centroid-based grouping with a heuristic priority rule. For each center route, it sums the prior selection counts of that route's customers and tries lower totals first. Initially, when no customer has been selected, it tries centers nearer the depot first.

The generated programs construct and prioritize their own delegations. Each returns an ordered list of groups of current route IDs, specifying both membership and priority. All methods use the same 10-route/200-customer delegation bounds, while only generated programs have the 64-delegation list cap. LKH or HGS constructs the repaired vehicle routes after a delegation is selected. The comparisons therefore include both candidate construction and selection under shared starting solutions, repair solvers and routing budgets.

We use the released checkpoints at training step 40,000 for both L2D baselines, matched to the repair solver as described in Section~\ref{sec:test-panels}.

\subsection{Instance design}
\label{app:instance-design}

Table~\ref{tab:panel-design} compares the instance designs used in this study. CVRP-X additionally varies depot type, customer demands, and vehicle capacity, and includes larger instances. The generator sets vehicle capacity to vary the typical number of customers per route.

For CVRP-S, we follow L2D's customer-location generator~\citep{li2021learning}. Uniform instances draw customer locations uniformly from the unit square. For clustered instances, the generator places three, five, or seven cluster centers and draws nearby locations using Gaussian offsets with standard deviation 0.07 per coordinate. The centers count as customers, and draws outside the square are discarded. Mixed instances combine clustered locations with uniform draws for roughly half of the customers.

\begin{figure}[!htbp]
 \centering
 \includegraphics[width=\linewidth]{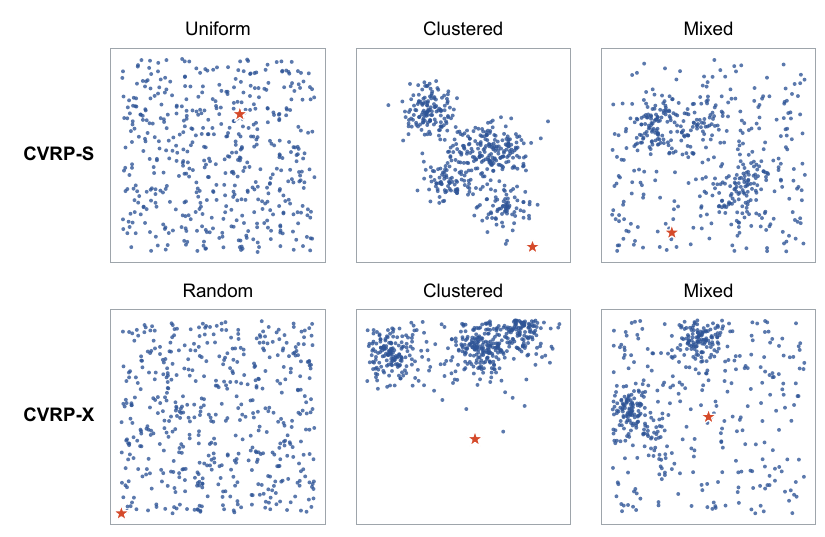}
 \caption{Six saved 500-customer test instances, with stars marking depots.}
 \label{fig:instance-layouts}
\end{figure}

For CVRP-X, the CVRPLIB generator draws random customer locations from a grid. Its clustered layout favors locations near cluster seeds, and its mixed layout combines random and clustered locations~\citep{uchoa2017benchmark}. Figure~\ref{fig:instance-layouts} shows examples from both families.

\begin{table}[!htbp]
\centering
\small
\setlength{\tabcolsep}{4pt}
\renewcommand{\arraystretch}{1.12}
\caption{CVRP-S and generated CVRP-X instance designs used in this study.}
\label{tab:panel-design}
\begin{tabular}{>{\raggedright\arraybackslash}p{0.22\linewidth}>{\raggedright\arraybackslash}p{0.33\linewidth}>{\raggedright\arraybackslash}p{0.38\linewidth}}
\hline
Feature & CVRP-S & CVRP-X \\
\hline
Study role & Discovery, validation and source-distribution testing & Transfer testing only \\
Customers ($N$) & 500 & 500, 1000 and 2000 \\
Test instances & 270 at $N=500$ & 90 at each size \\
Customer distribution & Uniform, clustered and mixed; clustered components use 3, 5 or 7 clusters & Random, clustered and mixed from the X-family generator \\
Depot-type groups & Not a separate design factor & Random, central and corner \\
Customer demands & Integers 1--9 & Seven regimes spanning unit, uniform, quadrant-dependent and many-small/few-large demands \\
Capacity & Fixed at 50 & Derived from six approximate route-size categories, from very short to ultra-long \\
\hline
\end{tabular}
\end{table}

Each CVRP-X size panel contains 30 instances per customer distribution and 30 per depot type (random, central, or corner), with ten instances for each combination. Demands include unit, uniform 1--10, uniform 5--10, uniform 1--100, uniform 50--100, quadrant-dependent and many-small/few-large regimes. The uniform 50--100 regime has 12 cases and each other regime has 13.
The panel has six categories for the approximate number of customers per route, with 15 cases in each category.
The design covers all individual levels and pairwise combinations, but only 90 of the 378 full categorical combinations. Each configuration has one generated instance, so the panel does not estimate variability within a fixed configuration. Coordinates are divided by the fixed 1000-unit grid extent before continuous-cost evaluation.

\subsection{Cost and time conventions}

Route cost is recomputed as the sum of continuous Euclidean edge lengths in the recorded coordinates.
Feasibility checks verify capacity and customer coverage. The elapsed routing clock includes startup, program execution, supervision, native solving, merging and checks. It excludes constructing the common starting solution. Full CVRP-S initialization times are saved. CVRP-X records contain only the ten sector-level LKH subprocess durations, excluding deterministic construction, serialization, parsing and validation. We therefore do not report end-to-end speedups that include initialization on CVRP-X.

Routing budgets are 120 seconds for CVRP-S and 300/600/1200 seconds for CVRP-X by size. If a run ends normally before its budget, its last checked solution is retained through the remaining time.
Curves use the latest checked solution at each plotted time, without interpolation between unknown arrival times. The progress figures display means without confidence bands.

For policy 135, 218 evaluations affected by interruptions or failures were restarted from their supplied starting solutions for the full routing budget; completed evaluations were retained.

\subsection{Confidence intervals}

We draw test instances with replacement and recompute the mean paired advantage, repeating this 20,000 times separately within each test panel. Each sampled instance retains its outcomes for all methods and three solver seeds.
Each CVRP-S resample contains 270 draws: 90 uniform and 30 from each of the three clustered and three mixed groups, defined by three, five or seven clusters.
Each CVRP-X resample contains 90 draws from its own size/solver panel. We use a fixed random seed (20261029) for reproducibility.

Each policy has its own family of 21 comparisons: three baselines in seven test panels (one CVRP-S panel and six CVRP-X size/solver panels). For each comparison, we take the 0.119th and 99.881st percentiles of its bootstrap mean advantages (exact tails $0.025/21$ and $1-0.025/21$). This gives a central 99.762\% interval for each comparison. Bonferroni correction targets nominal 95\% simultaneous coverage across all 21 comparisons for that policy.

Intervals describe instance variability for the retained programs and fixed solver seeds. Validation selected policy 135 before testing. Policies 143 and 147 were added as descriptive exploratory comparisons after transfer inspection, using their unchanged validation ranks.

\FloatBarrier

\section{Whole-instance solver comparisons}
\label{app:reference-evidence}

Whole-instance LKH and HGS runs provide secondary quality references. They do not enter prompts, fitness or selection. LKH and HGS each have 1080 seed runs on 540 instances: 270 CVRP-S cases and 90 CVRP-X cases per size.

For a reference solver $q\in\{\mathrm{LKH},\mathrm{HGS}\}$, let $C_{iq,s}(H_q)$ be its final cost on instance $i$ with seed $s$ at the reference budget $H_q$. The empirical reference averages final costs over two solver seeds. A comparison method $a$ has seed-averaged cost $\bar C_{ia}(T)$ at routing budget $T$. We compute its percentage gap to the reference on each instance and average over the $n$ instances in the test panel.
\begin{equation}
 \begin{aligned}
 R_{iq}&=\tfrac12[C_{iq,100}(H_q)+C_{iq,101}(H_q)],\\
 g_{ia\mid q}(T)&=100\frac{\bar C_{ia}(T)-R_{iq}}{R_{iq}},\qquad
 \bar g_{a\mid q}(T)=\frac{1}{n}\sum_{i=1}^{n}g_{ia\mid q}(T).
 \end{aligned}
 \label{eq:metric-reference}
\end{equation}
Lower values in~\eqref{eq:metric-reference} are better; zero matches the empirical reference cost. Normalization precedes instance averaging. These empirical references do not certify optimality.

Whole-instance reference runs receive 1800, 3600 and 7200 seconds for $N=500,1000,2000$, respectively.
Each select-solve-merge routing run receives 120 seconds on CVRP-S and 300, 600 and 1200 seconds on CVRP-X instances with $N=500,1000,2000$, respectively. Whole-instance LKH receives the supplied starting solution, and timing starts after the input file has been written. Recorded solution times include the delay between the solver producing output and the monitoring process detecting it. Whole-instance HGS starts without a supplied initial solution. Its progress is recorded using elapsed wall-clock time rather than the solver's internal CPU timestamps. These differences preclude matched-budget speedup comparisons.

Tables~\ref{tab:canonical_reference_gaps} and~\ref{tab:canonical-hgs-reference-gaps} give descriptive means.

\begin{table}[!htbp]
\centering\begingroup\setlength{\tabcolsep}{3pt}\scriptsize
\caption{Gap to the whole-instance LKH reference (\%).}
\label{tab:canonical_reference_gaps}
\small
\begin{tabular}{llrrrrrr}
\hline
Panel & $n$ & 135 & 143 & 147 & Centroid & \LTwoDUniform{} & \LTwoDClusteredMixed{} \\
\hline
CVRP-S 500/LKH & 270 & -0.8080 & -0.8649 & -0.8343 & -0.8766 & -0.9618 & \textbf{-0.9804} \\
CVRP-X 500/LKH & 90 & 0.2149 & \textbf{0.0752} & 0.0964 & 0.1895 & 0.4796 & 0.7664 \\
CVRP-X 1000/LKH & 90 & 0.0070 & \textbf{-0.1405} & -0.1316 & 0.0654 & 0.4081 & 0.6458 \\
CVRP-X 2000/LKH & 90 & -0.2322 & -0.2771 & \textbf{-0.2891} & -0.2226 & 0.2719 & 0.4016 \\
CVRP-X 500/HGS & 90 & -0.1470 & \textbf{-0.2504} & -0.1847 & -0.2031 & -0.1654 & -0.0898 \\
CVRP-X 1000/HGS & 90 & -0.4371 & \textbf{-0.5679} & -0.5181 & -0.4552 & -0.3393 & -0.2436 \\
CVRP-X 2000/HGS & 90 & -0.6981 & \textbf{-0.7736} & -0.7371 & -0.6733 & -0.6113 & -0.4357 \\
\hline
\end{tabular}
\endgroup
\par\smallskip
\begin{minipage}{.98\linewidth}\footnotesize Lower is better; zero matches the empirical LKH reference. These descriptive means use the unequal routing/reference budgets specified in Appendix~\ref{app:reference-evidence}. Bold marks row minima.\end{minipage}
\end{table}

\begin{table}[!htbp]
\centering
\scriptsize
\setlength{\tabcolsep}{3 pt}
\caption{Gap to the whole-instance HGS reference (\%).}
\label{tab:canonical-hgs-reference-gaps}
\small
\begin{tabular}{lrrrrrr}
\hline
Panel & 135 & 143 & 147 & Centroid & \LTwoDUniform{} & \LTwoDClusteredMixed{} \\
\hline
CVRP-S 500/LKH & +1.1771 & +1.1189 & +1.1501 & +1.1066 & +1.0206 & \textbf{+1.0008} \\
CVRP-X 500/LKH & +1.2129 & \textbf{+1.0718} & +1.0932 & +1.1879 & +1.4807 & +1.7689 \\
CVRP-X 1000/LKH & +0.9653 & \textbf{+0.8163} & +0.8253 & +1.0259 & +1.3719 & +1.6105 \\
CVRP-X 2000/LKH & +0.5364 & +0.4911 & \textbf{+0.4791} & +0.5469 & +1.0453 & +1.1761 \\
CVRP-X 500/HGS & +0.8458 & \textbf{+0.7415} & +0.8082 & +0.7897 & +0.8277 & +0.9039 \\
CVRP-X 1000/HGS & +0.5154 & \textbf{+0.3829} & +0.4333 & +0.4983 & +0.6152 & +0.7115 \\
CVRP-X 2000/HGS & +0.0656 & \textbf{-0.0110} & +0.0259 & +0.0914 & +0.1536 & +0.3304 \\
\hline
\end{tabular}
\par\smallskip
\begin{minipage}{0.98\linewidth}\footnotesize Lower is better; negative values beat the empirical HGS reference. All cases are included. In 12 of the 1080 whole-instance HGS reference runs, HGS returned no feasible solution; the supplied feasible starting solution was retained when computing the reference costs. These are descriptive means without confidence intervals. Bold marks row minima.\end{minipage}
\end{table}

\FloatBarrier

\section{Discovered heuristics}
\label{app:heuristics}

We reproduce each policy's original discovery hypothesis, followed by a brief overview. 
The hypotheses refer to earlier proposals.
All three programs form groups of nearby routes using estimated savings from moving customers between them. They lower the priority of large groups and repeated selections. LKH or HGS performs the repairs.

\subsection{Policy 135: primary}
\label{hyp:policy135}
\begin{Verbatim}
Candidate 134 was 0.0531 pp worse at 120 seconds, while candidate 122’s pair-level triple rescue improved by only 0.0393 pp, below promotion. I replace witness-centered widening with cooperation-gated rescue: complete one-/two-customer swaps and relocations are evaluated net of each outgoing block’s standalone reinsertion gain, and only pairs without a positive residual witness receive three-customer search. These pairs and independent contact seeds grow five-, seven-, and nine-route patches, ordered by alternating saturated residual-transfer/contact priorities with exposure-diverse smaller backups. This aims to improve 120-second partner coverage, but the combined gate may suppress useful gains and local witnesses need not survive cold-start LKH.
\end{Verbatim}

Policy 135 estimates savings from moving or swapping short customer sequences between routes. It subtracts the savings obtainable by moving those same sequences within their original routes. It considers three-customer sequences only when one- and two-customer moves offer little additional saving.

\subsection{Policy 143: exploratory}
\label{hyp:policy143}
\begin{Verbatim}
Candidate 142 was 0.1495 pp worse at 60 seconds, with no 120-second measurement, so I test asymmetric transfer coverage and route-saturated ranking instead of summed witnesses. On a bounded contact graph, every singleton position is examined while two-/three-customer blocks are position-sampled; feasible swaps and relocations seed ordinary five-, seven-, and nine-route contact patches. Alternating transfer/contact priorities with exposure-diverse smaller backups aims to preserve slack-using partners without exhaustive longer-block enumeration and improve 120-second gain throughput. Complete singleton coverage may overvalue small local moves, and these witnesses need not survive cold-start LKH’s mandated vehicle count.
\end{Verbatim}

Policy 143 checks moves involving every individual customer and samples two- and three-customer sequences at evenly spaced positions. It uses the estimated savings directly, without subtracting savings from rearrangements within a route.

\subsection{Policy 147: exploratory}
\label{hyp:policy147}
\begin{Verbatim}
Candidate 146 was 0.0349 pp worse at 120 seconds, while candidate 143’s 0.0496 pp improvement remained below promotion. I replace farthest-first longer-block sampling with positional samples supplemented at cumulative-demand quartiles, retaining every singleton; feasible swaps and relocations seed five-, seven-, and nine-route contact patches. Alternating route-saturated transfer/contact priorities with size penalties and exposure-diverse smaller backups aims to improve 120-second partner coverage without growth branches. Demand landmarks may overemphasize heavy customers, and local transfer witnesses need not survive cold-start LKH’s mandated vehicle count.
\end{Verbatim}

Policy 147 also checks every individual customer and samples longer sequences by position. It adds samples near the points where demand accumulated along a route reaches 25\%, 50\% and 75\% of its total.

\subsection{Program interface and source}
\label{app:program-interface}

The exact evaluated source for policies 135, 143, and 147 is available in the following repository.
\begin{center}
    \href{https://github.com/MaxWangZY/revisiting-learning-to-delegate}{github.com/MaxWangZY/revisiting-learning-to-delegate}.
\end{center}
Each program implements
\begin{quote}
\small\texttt{choose\_delegations(problem, state, history, limits)}.
\end{quote}
The four arguments provide the instance data, current routes and selection statistics, completed repair history, and delegation limits. The program returns groups of current route IDs in priority order. It can return at most 64 delegations, with at most 10 routes and 200 customers per delegation.

\end{document}